\documentclass{article}

    \PassOptionsToPackage{numbers, compress}{natbib}

     \usepackage[preprint]{neurips_2021}

\usepackage[utf8]{inputenc} 
\usepackage[T1]{fontenc}    
\usepackage{hyperref}       
\usepackage{url}            
\usepackage{booktabs}       
\usepackage{amsfonts}       
\usepackage{nicefrac}       
\usepackage{microtype}      
\usepackage{amsmath, amssymb}
\usepackage{mathtools}
\usepackage{algorithm}
\usepackage[noend]{algpseudocode}
\usepackage{lipsum}
\usepackage{multirow}
\usepackage[dvipsnames]{xcolor}
\usepackage[export]{adjustbox}
\usepackage[title]{appendix}
\PassOptionsToPackage{usenames,dvipsnames}{xcolor}
\usepackage{cleveref}
\newcommand\myshade{85}
\colorlet{mylinkcolor}{violet}
\colorlet{mycitecolor}{YellowOrange}
\colorlet{myurlcolor}{Aquamarine}

\hypersetup{
  linkcolor  = mylinkcolor!\myshade!black,
  citecolor  = mycitecolor!\myshade!black,
  urlcolor   = myurlcolor!\myshade!black,
  colorlinks = true,
}

\title{Contrastive Mixup: Self- and Semi-Supervised learning for Tabular Domain  }

\author{%
  Sajad Darabi \\
  UCLA \\
  \texttt{sajad.darabi@cs.ucla.edu} \\ \And
  Shayan Fazeli \\
  UCLA \\
  \texttt{shayan.fazeli@cs.ucla.edu} \\\And
  Ali Pazokitoroudi \\
  UCLA \\
  \texttt{alipazoki@cs.ucla.edu} \\ \And
    Sriram Sankararaman \\
  UCLA \\
  \texttt{sriram@cs.ucla.edu} \\ \And
    Majid Sarrafzadeh \\
  UCLA \\
  \texttt{majid@cs.ucla.edu} \\
  
}

\begin{document}

\maketitle

\begin{abstract}
Recent literature in self-supervised has demonstrated significant progress in closing the gap between supervised and unsupervised methods in the image and text domains. These methods rely on domain-specific augmentations that are not directly amenable to the tabular domain. Instead, we introduce Contrastive Mixup, a semi-supervised learning framework for tabular data and demonstrate its effectiveness in limited annotated data settings. Our proposed method leverages Mixup-based augmentation under the manifold assumption by mapping samples to a low dimensional latent space and encourage interpolated samples to have high a similarity within the same labeled class. Unlabeled samples are additionally employed via a transductive label propagation method to further enrich the set of similar and dissimilar pairs that can be used in the contrastive loss term. We demonstrate the effectiveness of the proposed framework on public tabular datasets and real-world clinical datasets.

\end{abstract}

\section{Introduction}

Deep learning has shown tremendous success in domains where large annotated datasets are readily available such as vision, text, speech via supervised learning. Implicitly learned by these models is an intermediate representation that lends itself useful for downstream tasks. Unfortunately, in many settings such as healthcare, large annotated datasets are not readily available to enable learning such valuable representations. As a result, there has been a push towards learning these in an unsupervised or semi-supervised manner as unannotated data on the other hand may be readily available for free and a lot of it in many cases. Recent literature has shown significant progress towards learning these useful representations without human-annotated data, closing the gap between supervised and unsupervised learning, and in some cases demonstrating superior transfer learning properties compared to its supervised counterpart \cite{he2020momentum, chen2020simple}.

Self-supervised methods have emerged as a promising approach to achieving appealing results in various applications without requiring labeled examples. This is typically done via pretext tasks closely related to the downstream tasks of interest and typically differs from domain to domain. For example, in the image domain colorization \cite{zhang2016colorful}, jigsaw puzzle\cite{sajjadi2016regularization}, rotation prediction \cite{gidaris2018unsupervised} have been previously presented as pretext tasks useful for learning such representations. Similarly, in the text domain, commonly used pretext tasks, such as predicting masked words and context words, have been widely used \cite{mikolov2013distributed, devlin2018bert}. More recently, contrastive learning methods introduced leverage domain specific transformations to create multiple semantically similar examples such as random cropping or flipping for images that preserve the semantic meaning and encourage the network to be invariant to such transformations achieving great success. Such pretext tasks and transformations cannot be readily applied that do not have the same structural information, as an example tabular data\footnote{Tabular data contains a set of rows (examples) and columns (features) that may be permutation invariant.}.

It is not clear how to generate new semantically similar examples for tabular data. Moreover, in many settings, tabular data contains both categorical and continuous features which require different treatment. In this work, we focus on tabular data settings that contain a small set of annotated samples and a relatively sizeable unlabeled set of samples. Specifically, we propose a framework for improving downstream task performance in this semi-supervised setting. Our method consists of a semi-self-supervised pretraining step where a feature reconstruction pretext task and a supervised contrastive loss term are used. Various forms of Mixup augmentation \cite{zhang2017mixup} has been used in the image domain, where new examples are created by taking convex combinations of pairs of examples. This may lead to low probable samples in the dataspace for tabular data. Instead, we leverage the manifold assumption \footnote{\textit{Manifold Assumption}: High-dimensional data lies (roughly) on a low-dimensional manifold.} and mix samples in the latent space to create multiple views for our contrastive loss term. The unlabeled subset is further leveraged by pre-training the encoder and using label propagation \cite{iscen2019label} to generate pseudo-labels for the unlabeled samples. Subsequently, the trained encoder and samples, for which we have generated pseudo-labels for, are transferred to a downstream task where a simple predictor with Mixup \cite{zhang2017mixup} augmentation is trained. We show that our proposed framework leads to improvements on various tabular datasets, such as UCI and Genomics (UK Biobank).

\section{Related Works}
Our work fits the semi-supervised learning framework \cite{chapelle2009semi} where both labeled and unlabeled samples are used to improve downstream task performance. We draw from recent literature in self-supervised representation learning, pseudo-labeling \cite{Lee_pseudo-label:the} and Mixup based supervised learning \cite{zhang2017mixup}.

At the core of the self-supervised methods are pre-text tasks, where labels are created from the raw unlabeled data itself, and supervised losses are then used to learn useful representations for downstream prediction tasks. In these lines of works, examples of domain-specific pre-text tasks such as jigsaw puzzle \cite{noroozi2016unsupervised}, colorization \cite{zhang2016colorful}, relative positioning prediction \cite{doersch2015unsupervised} have been introduced for images,  masked word prediction \cite{mikolov2013distributed, mnih2008scalable}, next sentence prediction \cite{logeswaran2018efficient} for text. There is also existing work on self-supervised/semi-supervised learning methods. For example, a similar in-painting task \cite{pathak2016context} can be used to predict masked features in a row as done in \cite{yin2020tabert, arik2019tabnet}. On the other hand, many recent self-supervised methods are based on contrastive representation learning \cite{chen2020simple}, in which domain-specific augmentation (e.g., random crop, random color distortion for images) are used to create "similar" samples, and the normalized cross-entropy loss \cite{sohn2016improved, oord2018representation} is used to increase the similarity of "positive" pairs in the latent space, and decrease the similarity of "negative" pairs. A downside of generating negative and positives without label information is that examples belonging to the same class may be pushed apart. In \cite{khosla2020supervised} authors leverage label information to consider many "similar" examples to be pulled closer to one another and farther away from the dissimilar examples. As these methods leverage properties inherent to the raw data, they are not amenable to the tabular domain, which is the focus of this work.

The setting we are considering fits the semi-supervised learning framework. Prior work on semi-supervised learning can be broadly separated into two main categories: methods that add an unsupervised loss term to the supervised task as a regularizer, e.g., \cite{sajjadi2016regularization, tarvainen2017mean, grandvalet2005semi} and methods that assign pseudo labels \cite{Lee_pseudo-label:the} to the unlabeled samples. Recently, \cite{yoon2020vime} proposed VIME, a state-of-the-art semi-supervised method for the tabular domain where they leverage consistency regularization and in-painting \cite{pathak2016context} inspired augmentation. In \cite{Lee_pseudo-label:the} the current network trained on labeled samples is used to infer pseudo-labels on unlabeled samples using a confidence threshold, which is then treated similar to labeled samples to minimize entropy. Transductive learning is more generic in that instead of training a generic classifier, the goal is to used patterns in the labeled set to infer labels for the unlabeled set. Label propagation has been widely used in transductive learning in the image domain in an online fashion where CNN features are used for few-shot learning \cite{douze2018low}. Along this line of work, recently \cite{iscen2019label} use label propagation in an offline fashion by treating the labeled and unlabeled samples as a bipartite graph where edges computed via diffusion similarity \cite{zhou2004learning}. In this work, we propose a semi-supervised framework for the tabular domain where we leverage Mixup \cite{zhang2017mixup} based augmentation, which interpolates samples using a convex combination and assigns soft labels according to the mixing ratio in the latent space \cite{verma2019manifold} and encourage samples interpolated from the same class to have high similarity.

\section{Preliminaries}
To present our method we formulate the self-supervised and semi-supervised problem. Consider a dataset with $N$ examples: Our assumption is that there is a small subset of this dataset for which labels are available: $\mathcal{D}_L = \{ (x_i, y_i)\}_{i=1}^{N_L}$, and the rest of the dataset is unlabeled: $\mathcal{D}_U = \{ (x_i)\}_{i=1}^{N_U}$  where $x_i$ are observations sampled from a data-generating distribution $p(x)$ and $y_i \in \{0, 1, \cdots, c\}$ is a discrete label set. We consider settings where the majority of the data is unlabeled i.e. $|\mathcal{D}_L| \ll |\mathcal{D}_U|$. In supervised learning a classifier $f : \mathcal{X} \rightarrow \mathcal{Y} \in \mathcal{F}$ is a function learned by an ML algorithm which aims at optimizing $f$ for a given loss function $l_A(\cdot)$ i.e. $f = \min_{f\in \mathcal{F}}\sum_{i=1}^N{l_A(f(x_i),y_i)}$. In this limited labeled data regime a supervised model is most likely to overfit, hence we propose to use the unlabeled samples to improve the models generalization.

\subsection{Self-Supervised Learning}
Self-supervised methods leverage unlabeled data to learn useful representations for downstream prediction tasks. Many techniques have been proposed for images where useful visual representations are learned through pre-text tasks such as in-painting, rotation, jig-saw \cite{noroozi2016unsupervised, pathak2016context, gidaris2018unsupervised}, and more recently, the gap between supervised and unsupervised models have drastically been reduced through contrastive representation learning method \cite{he2020momentum, chen2020simple}. Generally, in contrastive representation, learning a batch of $N$ samples is augmented through an augmentation function $\text{Aug}(.)$ to create a multi-viewed batch with $2N$ pairs, $\{\tilde{x_i}, \tilde{y_i}\}_{i=1\cdots2N}$ where $\tilde{x}_{2k}$ and $\tilde{x}_{2k-1}$ are two random augmentations of the same sample $x_k$ for $k = \{1, \cdots, N\}$. The samples are fed to an encoder $e: x \rightarrow z$ which takes a sample $x \in \mathcal{X}$, to obtain a  latent representation $z = e(x)$.  Typically when defining a pre-text task, a predictive model is trained jointly to minimize a self-supervised loss function $l_{ss}$.
 
 \begin{equation}
     \min\limits_{e, h} \mathbb{E}_{(x, \tilde{y}) \sim P(X, \tilde{Y})} \big[l(\tilde{y},h \circ e (x)] 
 \end{equation}
 
 where $h$  maps $z$ to an embedding space $h: z \rightarrow v$. Within a mutliviewed batch, $i \in \mathcal{I} = \{1, \cdots 2N\}$ the self supervised loss is defined as

 \begin{equation}
    \label{eqn:closs}
    l = \sum_{i\in \mathcal{I}} - \text{log} \Big(\frac{\text{exp}(\text{sim}(v_i, v_{j(i)})/\tau)}{\sum_{n \in  \mathcal{I} \backslash \{i\}}{\text{exp}(\text{sim}(v_i, v_{n})/\tau)}}\Big)
 \end{equation}
 
 Here, $\text{sim}(\cdot , \cdot) \in \Re^+$ is a similarity function (e.g. dot product or cosine similarity), $\tau \in \Re^+$ is a scalar temperature parameter, $i$ is the \textit{anchor}, $\mathcal{A}(i)$ is the \textit{positive(s)} and $\mathcal{I} \backslash \{i\}$ are the \textit{negatives}. The positive and negative samples refer to samples that are semantically similar and dissimilar respectively. Intuitively, the objective of this function is to bring the positives and the anchor closer in the embedding space $v$ than the anchor and the negatives, i.e. $\text{sim}(v^{a}, v^{+}) > \text{sim}(v^{a}, v^{-})$, where $v^a$ is the anchor and $v^+$, $v^-$ are the positive and negative respectively.
 
\subsection{Semi-Supervised Learning}

In semi-supervised learning (SSL), the dataset is comprised of two disjoint sets $D_L$. $D_U$, where predictive model $f$ is optimized to minimize a supervised loss, jointly with an unsupervised loss. In other words:
\begin{equation}
    \min\limits_{f} \mathbb{E}_{(x,y) \sim P(X, Y)}\big[l(y, f(x))\big] + \beta  \mathbb{E}_{(x,y_{ps}) \sim P(X, Y_{ps})}\big[l_u(y_{ps}, f(x))\big]
\end{equation}

The first term is estimated over the small labeled subset $\mathcal{D}_U$, and the second unsupervised loss is estimated over the more significant unlabeled subset. The unsupervised loss function $l_u$ is defined to help the downstream prediction task, such as consistency loss training \cite{noroozi2016unsupervised,tarvainen2017mean}, or in our case, a supervised objective on pseudo-labeled samples \cite{Lee_pseudo-label:the}.

\section{Method}
This section describes our proposed method Contrative Mixup, a semi-supervised method for multi-modal tabular data where structural (spatial or sequential) data augmentations are not readily available. To this end, we first propose our semi-supervised training to learn an encoder and subsequently propose to train a classifier using the pre-trained encoder and pseudo-labels.

\subsection{Semi-Self-Supervised Learning for Tabular Data}
\begin{figure}
    \centering
    \includegraphics{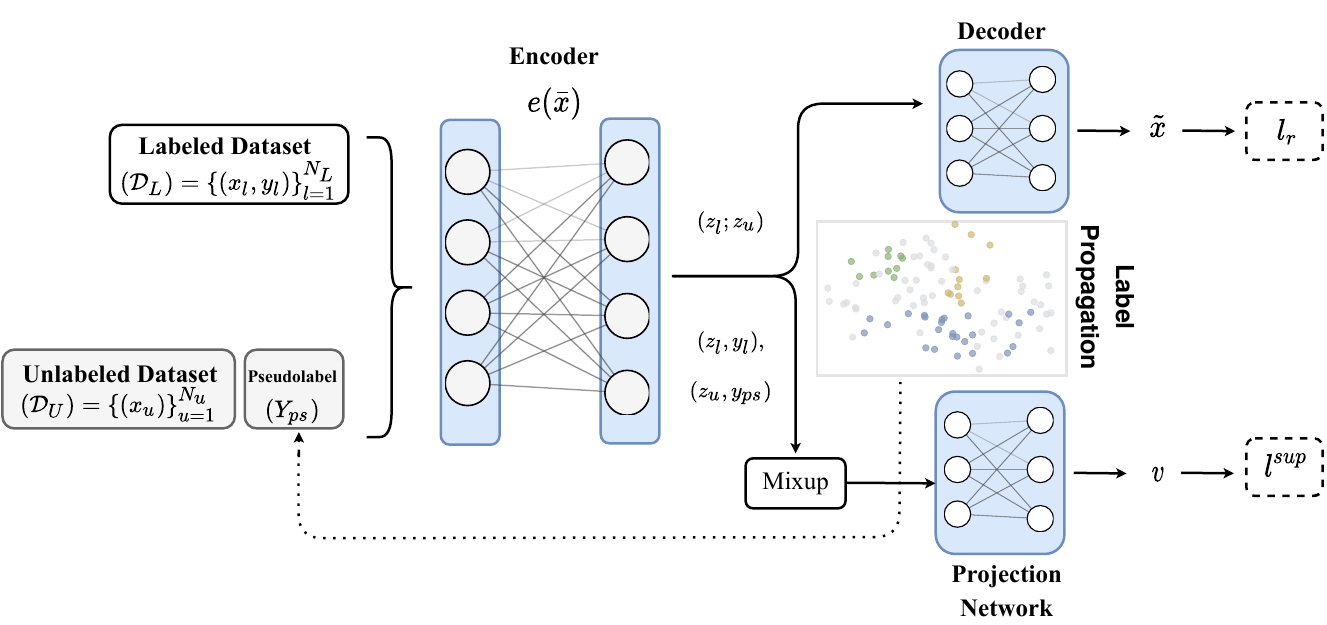}
    \caption{Overview of our semi-self-supervised framework. The encoder is trained using both labeled and unlabeled subsets via the reconstruction loss and contrastive loss terms. Pseudo-labeles are used }
    \label{fig:overview}
\end{figure}
We make use of the manifold assumption where high dimensional data roughly lie on a low dimensional manifold and then leverage Mixup \cite{zhang2017mixup} based data interpolation for creating positive and negative samples. By doing so we mitigate creating low-probable samples in the original data space.

In our setting we represent the mutli-modal tabular data rows $x_i$ as a concatenation of discrete $D = [D_1, \cdots, D_{|D|}]$ and continuous features $\mathcal{C} = [C_1, \cdots, C_{|\mathcal{C}|}]$. The raw features $x_i \in \Re^d$ are fed through an embedding layer $E: x \rightarrow \bar{x}$ that results in a feature vector $\bar{x} \in \Re^{|C| + \sum_i^{|D|}{d_{|\mathcal{D}_i|}}}$, that is a concatenation of the continuous features $\mathcal{C}$ and embedded discrete features $\mathcal{D}$, where $d_{|D_i|}$ is the embedding dimension for each discrete feature $\mathcal{D}_i$. The embedded features are fed to an encoder $z = e(\bar{x})$, and subsequently fed to a feature estimation pre-text task, as well as a semi-supervised contastive loss term shown in Figure \ref{fig:overview}.

In the tabular domain, data augmentation commonly used in the image domain cannot be used. Instead, we propose to interpolate between samples of the same class to create positive examples and use a supervised contrastive loss term in the latent space. Given a batch of labeled examples $\mathcal{D_{B}} = \{x_k, y_k\}^K_{k=1}$, we create a new labeled sample $(\hat{x}, \hat{y})$ by interpolating within the same labeled pair of examples
\begin{equation}
    \hat{x} = \lambda x_1 + (1 - \lambda) x_2
\end{equation}

where $\lambda$ is a scalar sampled from a random uniform $\lambda \sim \mathcal{U}(0, \alpha)$ with $\alpha \in [0, 0.5]$. The newly generated sample $\hat{x}$ will be $\lambda$ close to $x_1$ and $1-\lambda$ to $x_2$ with the same label as $x_1$ and $x_2$, i.e. $y_1=y_2=\hat{y}$. As opposed to randomly interpolating between samples and enforcing closeness between samples of different labels, we encourage samples of the same label to lie close to one another in the latent space.

Applying Mixup in the input space for tabular data may lead to low probable samples due to the multi-modality of the data and presence of categorical columns. Instead, we map samples to the hidden space and interpolate there. More concretely, given an encoder $e$, that is comprised of $T$ layers $f_t$, for $t \in \{1, \cdots T\}$. The samples are fed through to an intermediate representation $h_t$ at layer $t$. This layer contains a more abstract representations of the input samples $x_1$ and $x_2$. The samples are interpolated within this intermediate layer as
\begin{equation}
    \tilde{h}^{t}_{12} = \lambda h^{t}_{1} + (1 - \lambda) h^{t}_{2} 
\end{equation}
where $h^{t}_i$ is obtained by feeding samples $\bar{x}_i$ through the encoder until layer $t$. Subsequently, the newly generated samples $\tilde{h}^t_{i'i}$ as well as the original samples $h^{t}_i$ are fed through the rest of the encoder layers $t, \cdots, T$ to obtain the latent representation $z$. In this space we distinguish between $z_l$ and $z_u$, which are the latent representation of labeled and unlabeled samples respectively in. Note that initially we only consider the labeled portion for the contrastive term, i.e. $(z_l, y_l)$ in Figure. \ref{fig:overview}. We define the contrastive loss term to encourage samples created from pairs of the same class to have high similarity. It is common practice to introduce a separate predictive model to map the latent representations to an embedding space via a projection network $h^{proj}$ where the contrastive loss term is defined. We use supervised contrastive loss \cite{khosla2020supervised} for the labelled set $\mathcal{D}_L$ as our augmentation views are within a class. It generalizes Eqn. \ref{eqn:closs} to an arbitrary number of positive samples, due to the presence of labels and examples belonging to the same class are encouraged to have high similarity, making the loss term more sample efficient. 
\begin{equation}
    \label{eqn:supervised-contrastive-loss}
    l^{sup}_{\tau} = \sum_{i\in \mathcal{I}} \frac{-1 }{P(i)} \sum_{p \in P(i)} \text{log} \Big(\frac{\text{exp}(\text{sim}(h^{proj}_i, h^{proj}_{p})/\tau)}{\sum_{n \in Ne(i)} \text{exp}(\text{sim}(h^{proj}_i, h^{proj}_{n})/\tau)}\Big)
\end{equation}
In the above, $P(i) = \{ p | p \in \mathcal{A}(i), y_i = \tilde{y}_p \}$ is the set of indices of positives with the same label as example $i$, $|P(i)|$ is its cardinality, and $Ne(i) = \{ n | n \in \mathcal{I}, y_i \neq y_n\}$. This objective function will encourage mixed-uped labeled samples and anchors of the same sample to be close leading to a better cluster-able representation. In addition to the above loss term the encoder is trained to minimize the feature reconstruction loss via a decoder $f_\theta(\cdot)$ 
\begin{equation}
\label{eqn:reconstruction}
    l_r(x_i) = \frac{|C|}{d}\sum_c^{|C|} || f_\theta \circ e_\phi(x_i)^c - x_i^c||_2^2 + \frac{|D|}{d}{\sum_j^{|D|}\sum_o^{d_{D_j}}{ \mathbf{1}[x_i^d = o]\log(f_\theta \circ e_\phi(x_i)^o)}}
\end{equation}

The semi-self supervised objective function can then be written as
\begin{equation}
    L = \mathbb{E}_{(x,y) \sim \mathcal{D}_L}\big[l_{\tau}^{sup}(y, f(x))\big] + \beta  \mathbb{E}_{x \sim \mathcal{D}_U \cup \mathcal{D}_L}\big[l_r(x)\big]
\end{equation}
The encoder is trained using this loss term over $K$ epochs, to warm-start the representations in the latent space prior to pseudo-labeling and leveraging the unlabeled samples.
\subsection{Psuedo-labeling Unlabeled Samples}
Thus far, we have only used the labelled set $\mathcal{D}_L$ in the contrastive loss term $l^{sup}_\tau$. To make use of the unlabeled set using $\mathcal{D}_U$ we proposed to use label propagation \cite{iscen2019label, zhou2004learning} after $K$ epochs of training with the supervised contrastive loss term $L^{sup}$. Given the encoder trained on $\mathcal{D}_{L}$ for $K$ epochs, we map the small labelled set $\mathcal{D}_L$, and a subset of the unlabeled set $S_U \subset \mathcal{D}_U$ to the latent space $z$ and construct an affinity matrix $G$
\begin{equation}
    g_{ij} := \begin{cases}
                    \text{sim}(z_i, z_j)~ &\text{if}~ i \neq j ~ \text{and} ~z_j \in \text{NN}_k(i) \\
                    0 & \text{otherwise}
              \end{cases}
\end{equation}
where $\text{NN}_k(i)$ is the $k$ nearest neighbor of sample $z_i$ and $\text{sim}(\cdot, \cdot) \Re^{+}$ is a similarity measure, e.g. $z_i^Tz_j$. We then obtain pseudolabels for our unlabeled samples by computing the diffusion matrix $C$ and setting $\tilde{y}_i :=\text{arg}\max\limits_j c_{ij}$, where

$$(I - \alpha \mathcal{A}) C = Y$$
Similar to \cite{iscen2019label, zhu2005semi} we use conjugate method to solve linear equations to obtain $C$ to enable efficient computation of the pseudo-labels. Here $\mathcal{A} = D^{-1/2}WD^{-1/2}$ is the adjacency matrix, $W = G^T + G$ and $D := \text{diag}(W1_n)$ is the degree matrix. Once we've obtained the pseudo-labels for the unlabeled subset $S_U$, we train the encoder with unlabeled samples treating the generated labels as ground truth
\begin{equation}
        L = \mathbb{E}_{(x,y) \sim \mathcal{D}_L}\big[l^{sup}(y, f(x))\big] + \gamma  \mathbb{E}_{(x,y_{ps}) \sim  S_U)}\big[l^{sup}(y_{ps}, f(x))\big] + \beta  \mathbb{E}_{x \sim \mathcal{D}_U}\big[l_r(x)\big]
\end{equation}

The pseudo-labels are updated every $f$ epoch of training with the above loss term.

\subsection{Predictor}
\begin{figure}
    \centering
    \includegraphics{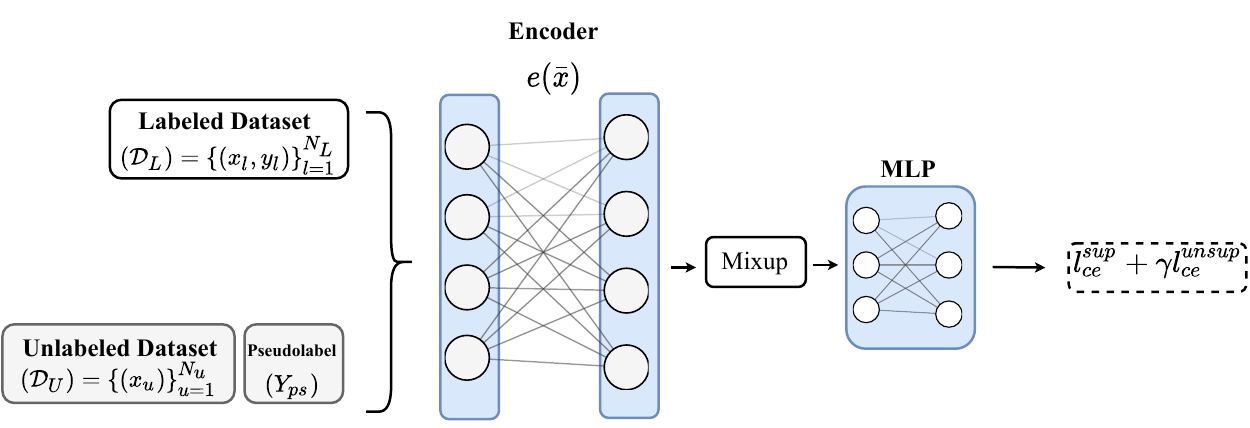}
    \caption{Overview transfering the semi-supervised pre-training steps to the downstream task. Encoder $e(\bar{x})$ is fixed and the predictor - multilayer perceptron (MLP) is trained using Mixup augmentation. $l_{ce}^{x}$ is the generic cross-entropy loss split into supervised (sup) for labeled subset and unsupervised (unsup) for the unlabeled subset.}
    \label{fig:predictor}
\end{figure}

Following the semi-supervised pre-training, the encoder is transferred to the downstream task along with the generated pseudo-labels to train the predictor on the downstream task. We leverage Mixup augmentation \cite{zhang2017mixup} in the latent space and feed samples to a set of fully connected layers as depicted in Figure \ref{fig:predictor}. 

\section{Experiment \& Emperical Results}
In this section we showcase the proposed framework on a set of different tabular datasets and application domains to demonstrate its effectiveness. We compare our semi-supervised framework with VIME \cite{yoon2020vime} another semi-supervised approach for the same problem set as a benchmark. To evaluate the pre-training phase, we compare with auto-encoder. We also compare with other semi-supervised method manifold Mixup \cite{verma2019manifold}. As a baseline, we include supervised methods, Logistic Regression, a 2-layer multi-layer perceptron network (MLP) that is used as the same architecture amongst other deep methods as the predictor network, and we also include CatBoost \cite{prokhorenkova2017catboost} as a gradient boosting tree method widely used on tabular data as it supports categorical columns. Additionally, we provide results for including various components of the proposed framework as ablation for the usefulness of each part of the method. In the experiments, self/semi-supervised use the labelled and unlabeled sets $\mathcal{D}_L$, $\mathcal{D}_U$ during training, and supervised models only used the labelled sets $\mathcal{D}_L$. We normalize the continuous columns to 0, 1 using Standard-scaler\footnote{\href{sklearn.preprocessing.StandardScaler}{https://scikit-learn.org/stable/modules/generated/sklearn.preprocessing.StandardScaler.html}}. We provide more details on the experimental setup in the Supplementary Material. The implementation of ContrastiveMixup can be found at \href{https://github.com/<anonmous>/ContrastiveMixup}{https://github.com/sajaddarabi/ContrastiveMixup}

\subsection{Public Tabular Datasets}
We compare the proposed method on four public UCI\footnote{\href{https://archive.ics.uci.edu/ml/datasets.php}{https://archive.ics.uci.edu/ml/datasets.php}} datasets: MNIST, where examples are interpreted as 784-dimensional feature vectors, UCI Adult, UCI Covertype, more details, are available in the supplementary. We use $10\%$ of the data as labelled and the rest as unlabeled; if the dataset contains an original test set, we use this to evaluate the methods; otherwise, we split the dataset $80\%$ train and $20\%$ test, and the ratios mentioned above follow. As we introduced embedding layers for categorical columns in our method, we choose the best of one-hot encoding categorical columns or embedding layers for other methods. The different variants of our methods for the ablation study are as follows:
\begin{itemize}
    \item \textbf{Supervised}: the pre-training is removed and only the predictor is trained (i.e. the same as MLP)
    \item \textbf{Self-SL only}: the pre-training consisting of labeled contrastive loss term and unsupervised reconstruction loss without pseudo-labeling. (i.e. $\gamma = 0)$
    \item \textbf{Self-SL + PL}: This is the pre-training with pseudo-labeling component added, without Mixup component when training the predictor.
\end{itemize}

\begin{table*}[h]%
\renewcommand{\arraystretch}{1.2}
\caption{\label{tab:public} Comparison on public tabular datasets.}
\begin{minipage}{\textwidth}
\begin{center}
\resizebox{\columnwidth}{!}{
\begin{tabular}{ll | ccccc}
\toprule
& &  \multicolumn{4}{c}{\textbf{Dataset}} \\
\textbf{Type} & \textbf{Method} & \textbf{MNIST} & \textbf{Adult} & \textbf{Blog Feedback} & \textbf{Covertype} \\
\hline
\multirow{3}{*}{\textbf{Supervised}}
& Logistic & {90.12} {\scriptsize($\pm 0.098$)}& {82.41} {\scriptsize($\pm 0.413$)}& {78.91} {\scriptsize($\pm 0.22$)}& {70.54} {\scriptsize($\pm 0.087$)}\\
& MLP & {93.69} {\scriptsize($\pm 0.234$)}& {83.19} {\scriptsize($\pm 0.663$)}& {79.63} {\scriptsize($\pm 0.519$)}& {75.95} {\scriptsize($\pm 0.202$)}\\
& CatBoost ($100\%$) & {97.41} {\scriptsize($\pm 0.098$)}& {87.54} {\scriptsize($\pm 0.075$)}& {85.08} {\scriptsize($\pm 0.088$)}& {88.64} {\scriptsize($\pm 0.077$)}\\ \hline

\multirow{3}{*}{\textbf{Semi-supervised}}
& AE & {94.72} {\scriptsize($\pm 0.127$)}& {84.18} {\scriptsize($\pm 0.078$)}& {80.09} {\scriptsize($\pm 0.199$)}& {79.67} {\scriptsize($\pm 0.296$)}\\
& Manifold Mixup & {94.92} {\scriptsize($\pm 0.012$)}& {84.68} {\scriptsize($\pm 0.279$)}& {80.24} {\scriptsize($\pm 0.652$)}& {78.79} {\scriptsize($\pm 0.135$)}\\
& VIME & {95.71} {\scriptsize($\pm 0.013$)}& {84.54} {\scriptsize($\pm 0.408$)}& {81.36} {\scriptsize($\pm 0.301$)}& {79.02} {\scriptsize($\pm 0.329$)}\\\hline
\multirow{4}{*}{\textbf{Ours (Ablation)}}
& Supervised & {93.69} {\scriptsize($\pm 0.234$)}& {83.19} {\scriptsize($\pm 0.663$)}& {79.63} {\scriptsize($\pm 0.519$)} & {75.95} {\scriptsize($\pm 0.202$)}\\
& Self-SL & {95.82} {\scriptsize($\pm 0.131$)}& {85.16} {\scriptsize($\pm 0.249$)}& {81.38} {\scriptsize($\pm 0.373$)}& {79.46} {\scriptsize($\pm 0.463$)}\\
& Self-SL+PL & {97.01} {\scriptsize($\pm 0.066$)}& {85.26} {\scriptsize($\pm 0.207$)}& {81.65} {\scriptsize($\pm 0.370$)}& {79.92} {\scriptsize($\pm 0.682$)}\\
&  Ours & \textbf{{97.58} {\scriptsize($\pm 0.078$)}}& \textbf{85.42} {\scriptsize($\pm 0.210$)}& \textbf{81.88} {\scriptsize($\pm 0.123$)}& \textbf{80.41} {\scriptsize($\pm 0.205$)}\\
\hline
\end{tabular}}
\end{center}
\end{minipage}
\end{table*}

From Table \ref{tab:public} we can see Contrastive Mixup consistently outperforms previous methods. Further, through our ablation we demonstrate the effectiveness of various components of our framework each provide benefit in improving downstream task performance. Pseudo-labeling consistently helps in improving performance, comparing Self-SL versus Self-SL+PL allbeit at varying degree for different datasets. As a reference Catboost trained on $100\%$ of the labeled samples is provided as well. Our results on $10\%$ labeled MNIST with the help of pseudo-labeling outperforms this reference point.

\subsection{Genomics Datasets}

We assessed the accuracy of our method on the UK Biobank \footnote{\href{http://www.ukbiobank.ac.uk}{http://www.ukbiobank.ac.uk} Application \# 33127} genotypes consisting of around 500,000 individuals genotyped at around 10 millions SNPs. In this experiment, we restricted our analysis to SNPs  with  minor allele frequency larger than $1\%$. Moreover, SNPs that fail the Hardy-Weinberg test at significance threshold $10^{-7}$  were removed. Our analysis is restricted to around 300,000 unrelated self-reported British white ancestry individuals. We selected four complex traits measured in UK Biobank. As including all of the SNPs in our analysis is computationally expensive, we therefore, for every trait select around 1000 significantly associated SNPs with smallest p-value based on a publicly available summary statistics\footnote{\href{ https://alkesgroup.broadinstitute.org/UKBB/}{ https://alkesgroup.broadinstitute.org/UKBB/}}. Note that the set of selected SNPs is different across traits after p-value filtering as each trait is associated with different genetic variants, hence every phenotype task could be considered a different dataset.

To explore the efficacy of our semi-supervised framework on limited labeled data sets in practical setting, we compared the accuracy of our method with state of the art methods by varying the number of labeled individuals and using the remaining individuals as unlabeled samples. The results on four phenotypes are shown in Figure \ref{fig:genomics_lr_acc_curve}. From these figures we can see the semi-supervised methods outperform logistic regression model for cases where we only have access to a few thousand labeled samples. For two out of the four phenotypes logistic regression model outperforms the deep supervised models when adequate labeled samples are available i.e. $> 10^4$. This may be due to only a subset of features being selected using $p$-value threshold and hence making deeper models prone to overfititng.
\begin{figure}
    \centering
    \begin{tabular}{ll}
    \includegraphics[width=.5\linewidth,valign=m]{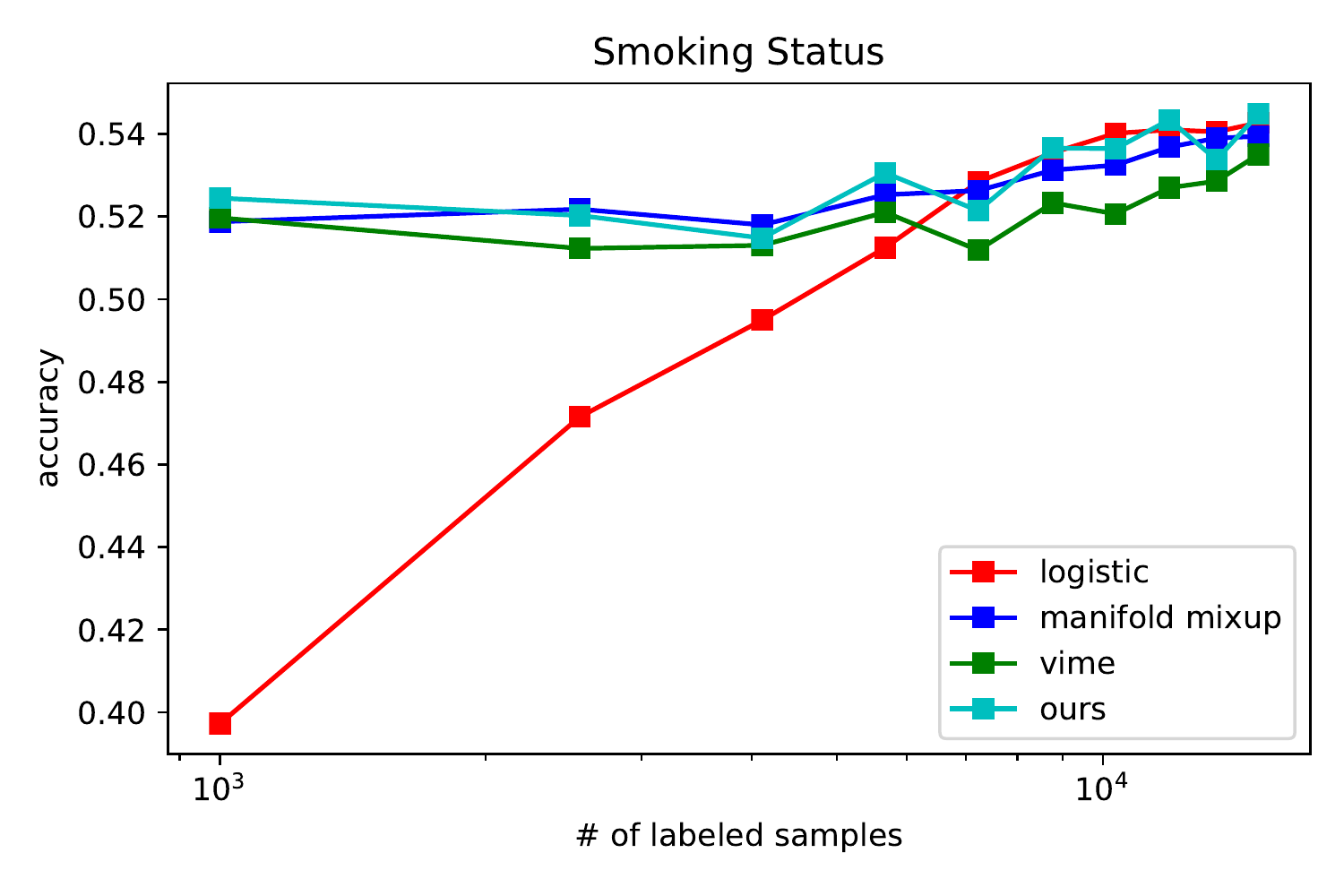} & \includegraphics[width=.5\linewidth,valign=m]{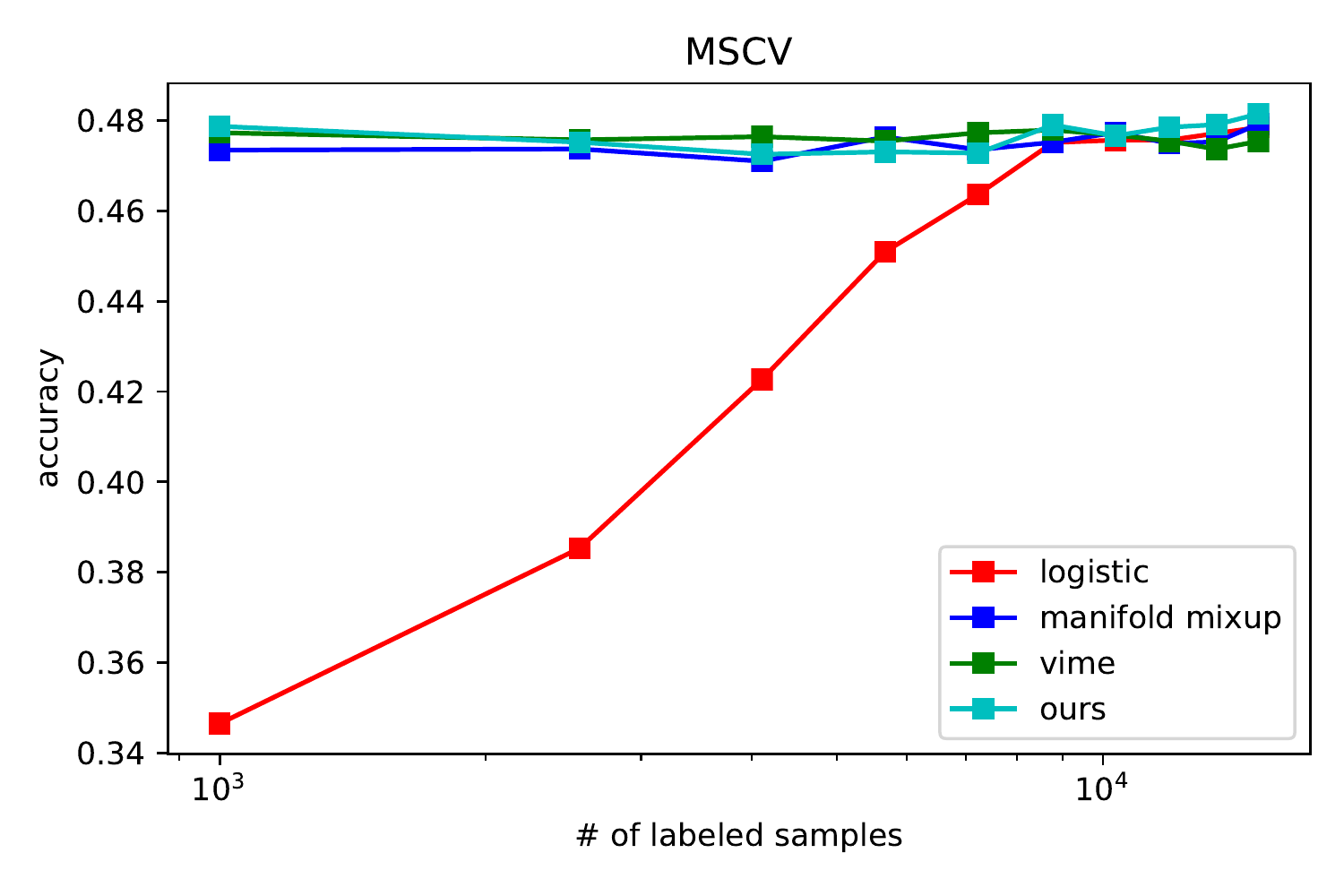}\\
    \includegraphics[width=.5\linewidth,valign=m]{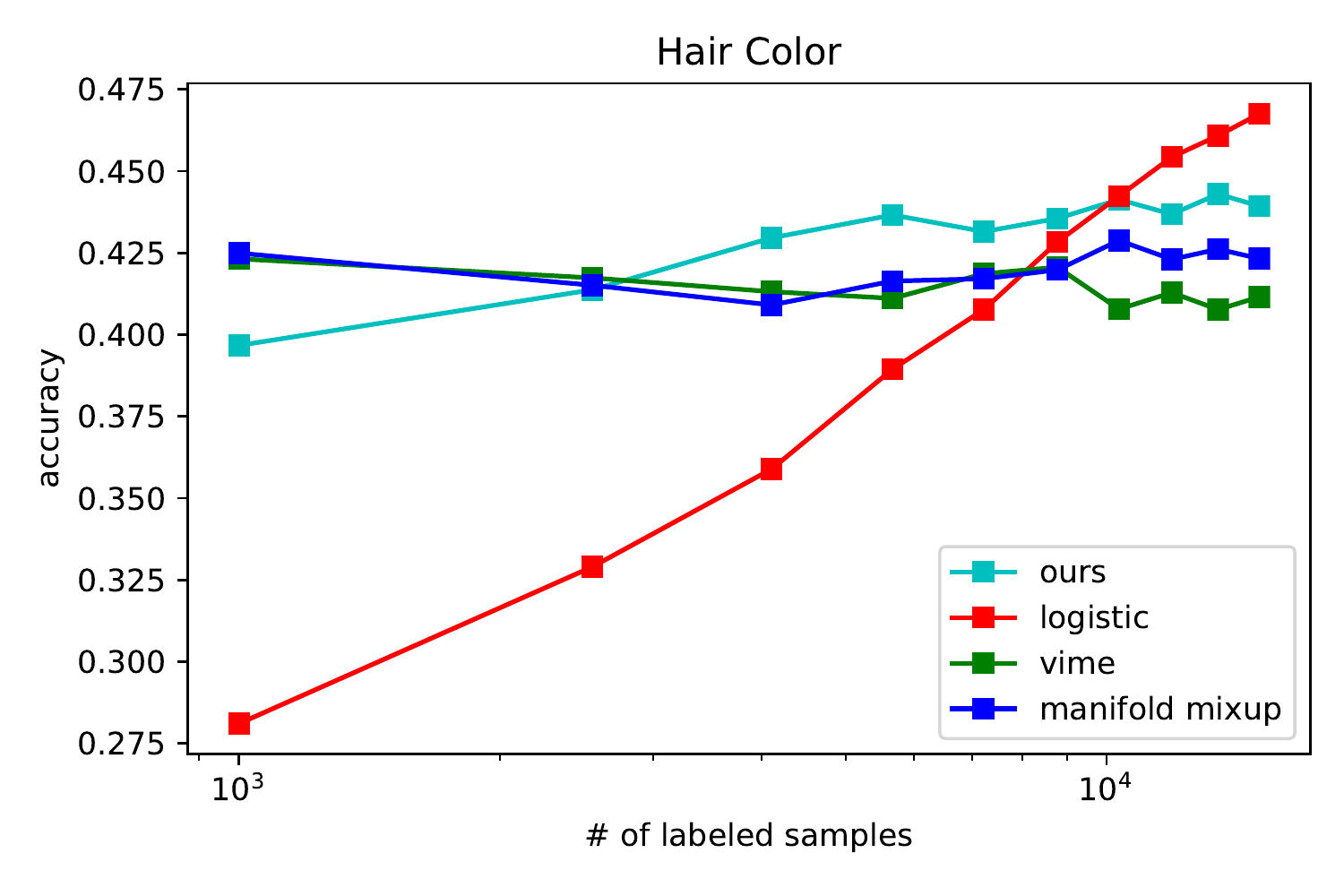} & \includegraphics[width=.5\linewidth,valign=m]{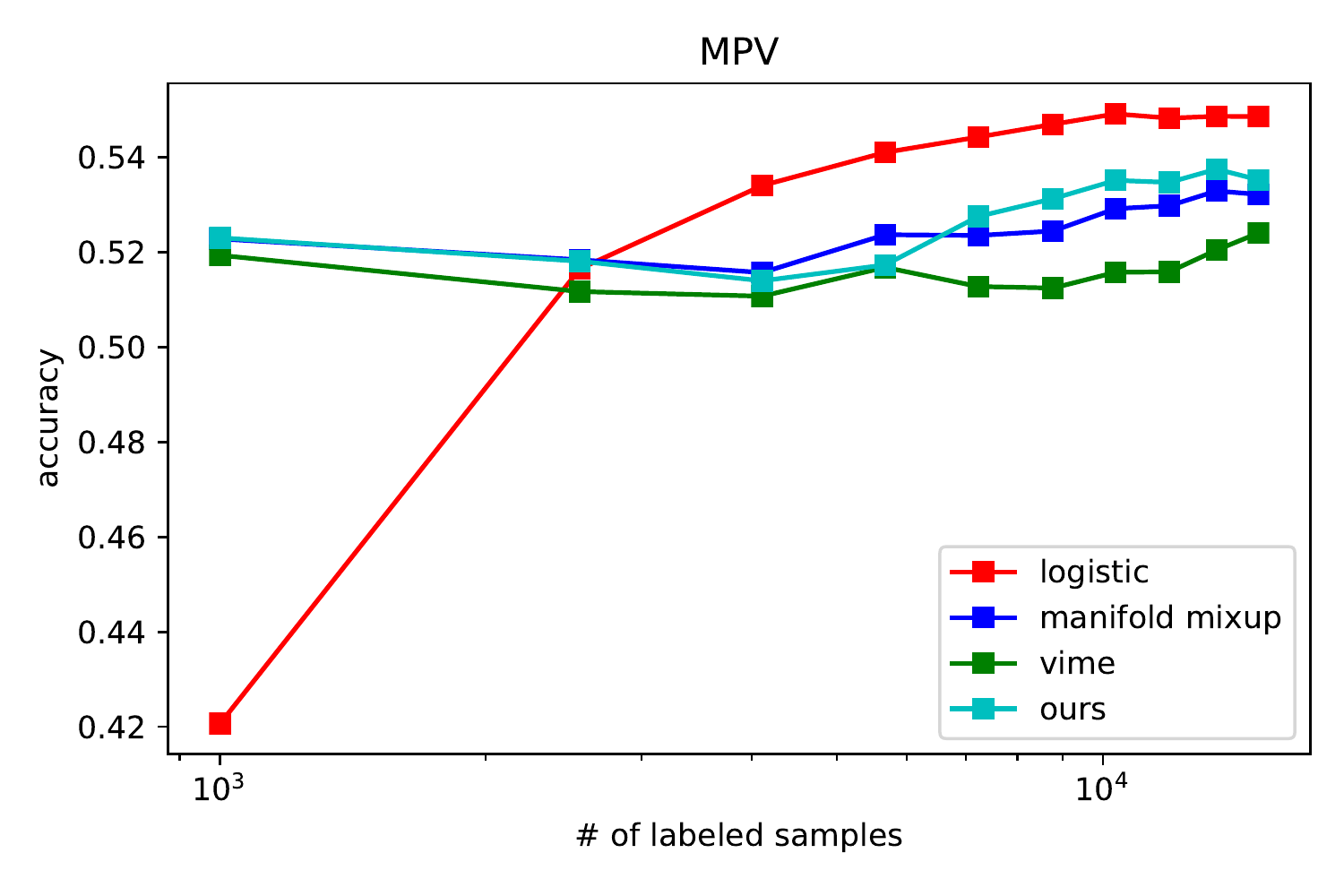}\\
    \end{tabular}
    \caption{Accuracy performance on four UK Biobank phenotypes across different number of labeled samples used as training set and the test set is fixed across experiments. The x axes is plotted in log-scale.}
    \label{fig:genomics_lr_acc_curve}
\end{figure}

\section{Conclusion}

Tabular data presents a different challenge compared to images and text, as similar structure or semantics aren't present, hence mitigating the transfer-ability of methods from those domain to the tabular domain. As a result extending semi-supervised methods that work well in those domains is more challenging for the tabular domain. Additionally, as most of the literature revolves around images and text not many pre-text tasks, and transformations have been investigated for such unstructured datasets where "correlations" or semantic meanings aren't immediately present in the data. Instead, we propose a framework for extending the recent contrastive learning paradigm to the tabular domain and help propel it's success in this domain as well. We do this by mapping samples to the latent space and creating new examples interpolating between samples in this space. We empirically show the effectiveness of the proposed method, and demonstrate how it improves learning from tabular data with limited labels. Further, improvements on pre-text tasks or augmentation methods for tabular datasets will dramatically improve the applicability of deep learning for these data modalities.

\section{Broader Impact}
Tabular data is very common in wide array of applications ranging from financial institutions, insurance companies, to health and clinical settings. These datasets contain both categorical and continuous features, such as demographic information, or real valued time series in finance datasets. As Deep Learning has shown great success in different data modalities such as text and images, by leveraging various pre-text tasks and domain specific augmentations for training in limited annotated data settings, by extending this over to the tabular domain many real-world applications will benefit and the applicability of Deep Learning will be greatly extended. The proposed method takes a step in this direction.

\bibliographystyle{plainnat}
\bibliography{bib}

\begin{thebibliography}{29}
\providecommand{\natexlab}[1]{#1}
\providecommand{\url}[1]{\texttt{#1}}
\expandafter\ifx\csname urlstyle\endcsname\relax
  \providecommand{\doi}[1]{doi: #1}\else
  \providecommand{\doi}{doi: \begingroup \urlstyle{rm}\Url}\fi

\bibitem[Arik and Pfister(2019)]{arik2019tabnet}
Sercan~O Arik and Tomas Pfister.
\newblock Tabnet: Attentive interpretable tabular learning.
\newblock \emph{arXiv preprint arXiv:1908.07442}, 2019.

\bibitem[Chapelle et~al.(2009)Chapelle, Scholkopf, and Zien]{chapelle2009semi}
Olivier Chapelle, Bernhard Scholkopf, and Alexander Zien.
\newblock Semi-supervised learning (chapelle, o. et al., eds.; 2006)[book
  reviews].
\newblock \emph{IEEE Transactions on Neural Networks}, 20\penalty0
  (3):\penalty0 542--542, 2009.

\bibitem[Chen et~al.(2020)Chen, Kornblith, Norouzi, and Hinton]{chen2020simple}
Ting Chen, Simon Kornblith, Mohammad Norouzi, and Geoffrey Hinton.
\newblock A simple framework for contrastive learning of visual
  representations, 2020.

\bibitem[Devlin et~al.(2018)Devlin, Chang, Lee, and Toutanova]{devlin2018bert}
Jacob Devlin, Ming-Wei Chang, Kenton Lee, and Kristina Toutanova.
\newblock Bert: Pre-training of deep bidirectional transformers for language
  understanding.
\newblock \emph{arXiv preprint arXiv:1810.04805}, 2018.

\bibitem[Doersch et~al.(2015)Doersch, Gupta, and
  Efros]{doersch2015unsupervised}
Carl Doersch, Abhinav Gupta, and Alexei~A Efros.
\newblock Unsupervised visual representation learning by context prediction.
\newblock In \emph{Proceedings of the IEEE international conference on computer
  vision}, pages 1422--1430, 2015.

\bibitem[Douze et~al.(2018)Douze, Szlam, Hariharan, and
  J{\'e}gou]{douze2018low}
Matthijs Douze, Arthur Szlam, Bharath Hariharan, and Herv{\'e} J{\'e}gou.
\newblock Low-shot learning with large-scale diffusion.
\newblock In \emph{Proceedings of the IEEE Conference on Computer Vision and
  Pattern Recognition}, pages 3349--3358, 2018.

\bibitem[Gidaris et~al.(2018)Gidaris, Singh, and
  Komodakis]{gidaris2018unsupervised}
Spyros Gidaris, Praveer Singh, and Nikos Komodakis.
\newblock Unsupervised representation learning by predicting image rotations.
\newblock \emph{arXiv preprint arXiv:1803.07728}, 2018.

\bibitem[Grandvalet et~al.(2005)Grandvalet, Bengio, et~al.]{grandvalet2005semi}
Yves Grandvalet, Yoshua Bengio, et~al.
\newblock Semi-supervised learning by entropy minimization.
\newblock In \emph{CAP}, pages 281--296, 2005.

\bibitem[He et~al.(2020)He, Fan, Wu, Xie, and Girshick]{he2020momentum}
Kaiming He, Haoqi Fan, Yuxin Wu, Saining Xie, and Ross Girshick.
\newblock Momentum contrast for unsupervised visual representation learning.
\newblock In \emph{Proceedings of the IEEE/CVF Conference on Computer Vision
  and Pattern Recognition}, pages 9729--9738, 2020.

\bibitem[hyun Lee()]{Lee_pseudo-label:the}
Dong hyun Lee.
\newblock Pseudo-label: The simple and efficient semi-supervised learning
  method for deep neural networks.

\bibitem[Iscen et~al.(2019)Iscen, Tolias, Avrithis, and Chum]{iscen2019label}
Ahmet Iscen, Giorgos Tolias, Yannis Avrithis, and Ondrej Chum.
\newblock Label propagation for deep semi-supervised learning.
\newblock In \emph{Proceedings of the IEEE/CVF Conference on Computer Vision
  and Pattern Recognition}, pages 5070--5079, 2019.

\bibitem[Khosla et~al.(2020)Khosla, Teterwak, Wang, Sarna, Tian, Isola,
  Maschinot, Liu, and Krishnan]{khosla2020supervised}
Prannay Khosla, Piotr Teterwak, Chen Wang, Aaron Sarna, Yonglong Tian, Phillip
  Isola, Aaron Maschinot, Ce~Liu, and Dilip Krishnan.
\newblock Supervised contrastive learning.
\newblock \emph{arXiv preprint arXiv:2004.11362}, 2020.

\bibitem[Logeswaran and Lee(2018)]{logeswaran2018efficient}
Lajanugen Logeswaran and Honglak Lee.
\newblock An efficient framework for learning sentence representations.
\newblock \emph{arXiv preprint arXiv:1803.02893}, 2018.

\bibitem[Mikolov et~al.(2013)Mikolov, Sutskever, Chen, Corrado, and
  Dean]{mikolov2013distributed}
Tomas Mikolov, Ilya Sutskever, Kai Chen, Greg Corrado, and Jeffrey Dean.
\newblock Distributed representations of words and phrases and their
  compositionality.
\newblock \emph{arXiv preprint arXiv:1310.4546}, 2013.

\bibitem[Mnih and Hinton(2008)]{mnih2008scalable}
Andriy Mnih and Geoffrey~E Hinton.
\newblock A scalable hierarchical distributed language model.
\newblock \emph{Advances in neural information processing systems},
  21:\penalty0 1081--1088, 2008.

\bibitem[Noroozi and Favaro(2016)]{noroozi2016unsupervised}
Mehdi Noroozi and Paolo Favaro.
\newblock Unsupervised learning of visual representations by solving jigsaw
  puzzles.
\newblock In \emph{European conference on computer vision}, pages 69--84.
  Springer, 2016.

\bibitem[Oord et~al.(2018)Oord, Li, and Vinyals]{oord2018representation}
Aaron van~den Oord, Yazhe Li, and Oriol Vinyals.
\newblock Representation learning with contrastive predictive coding.
\newblock \emph{arXiv preprint arXiv:1807.03748}, 2018.

\bibitem[Pathak et~al.(2016)Pathak, Krahenbuhl, Donahue, Darrell, and
  Efros]{pathak2016context}
Deepak Pathak, Philipp Krahenbuhl, Jeff Donahue, Trevor Darrell, and Alexei~A
  Efros.
\newblock Context encoders: Feature learning by inpainting.
\newblock In \emph{Proceedings of the IEEE conference on computer vision and
  pattern recognition}, pages 2536--2544, 2016.

\bibitem[Prokhorenkova et~al.(2017)Prokhorenkova, Gusev, Vorobev, Dorogush, and
  Gulin]{prokhorenkova2017catboost}
Liudmila Prokhorenkova, Gleb Gusev, Aleksandr Vorobev, Anna~Veronika Dorogush,
  and Andrey Gulin.
\newblock Catboost: unbiased boosting with categorical features.
\newblock \emph{arXiv preprint arXiv:1706.09516}, 2017.

\bibitem[Sajjadi et~al.(2016)Sajjadi, Javanmardi, and
  Tasdizen]{sajjadi2016regularization}
Mehdi Sajjadi, Mehran Javanmardi, and Tolga Tasdizen.
\newblock Regularization with stochastic transformations and perturbations for
  deep semi-supervised learning.
\newblock \emph{arXiv preprint arXiv:1606.04586}, 2016.

\bibitem[Sohn(2016)]{sohn2016improved}
Kihyuk Sohn.
\newblock Improved deep metric learning with multi-class n-pair loss objective.
\newblock In \emph{Proceedings of the 30th International Conference on Neural
  Information Processing Systems}, pages 1857--1865, 2016.

\bibitem[Tarvainen and Valpola(2017)]{tarvainen2017mean}
Antti Tarvainen and Harri Valpola.
\newblock Mean teachers are better role models: Weight-averaged consistency
  targets improve semi-supervised deep learning results.
\newblock \emph{arXiv preprint arXiv:1703.01780}, 2017.

\bibitem[Verma et~al.(2019)Verma, Lamb, Beckham, Najafi, Mitliagkas, Lopez-Paz,
  and Bengio]{verma2019manifold}
Vikas Verma, Alex Lamb, Christopher Beckham, Amir Najafi, Ioannis Mitliagkas,
  David Lopez-Paz, and Yoshua Bengio.
\newblock Manifold mixup: Better representations by interpolating hidden
  states.
\newblock In \emph{International Conference on Machine Learning}, pages
  6438--6447. PMLR, 2019.

\bibitem[Yin et~al.(2020)Yin, Neubig, Yih, and Riedel]{yin2020tabert}
Pengcheng Yin, Graham Neubig, Wen-tau Yih, and Sebastian Riedel.
\newblock Tabert: Pretraining for joint understanding of textual and tabular
  data.
\newblock \emph{arXiv preprint arXiv:2005.08314}, 2020.

\bibitem[Yoon et~al.(2020)Yoon, Zhang, Jordon, and van~der
  Schaar]{yoon2020vime}
Jinsung Yoon, Yao Zhang, James Jordon, and Mihaela van~der Schaar.
\newblock Vime: Extending the success of self-and semi-supervised learning to
  tabular domain.
\newblock \emph{Advances in Neural Information Processing Systems}, 33, 2020.

\bibitem[Zhang et~al.(2017)Zhang, Cisse, Dauphin, and
  Lopez-Paz]{zhang2017mixup}
Hongyi Zhang, Moustapha Cisse, Yann~N Dauphin, and David Lopez-Paz.
\newblock mixup: Beyond empirical risk minimization.
\newblock \emph{arXiv preprint arXiv:1710.09412}, 2017.

\bibitem[Zhang et~al.(2016)Zhang, Isola, and Efros]{zhang2016colorful}
Richard Zhang, Phillip Isola, and Alexei~A Efros.
\newblock Colorful image colorization.
\newblock In \emph{European conference on computer vision}, pages 649--666.
  Springer, 2016.

\bibitem[Zhou et~al.(2004)Zhou, Bousquet, Lal, Weston, and
  Sch{\"o}lkopf]{zhou2004learning}
Dengyong Zhou, Olivier Bousquet, Thomas~N Lal, Jason Weston, and Bernhard
  Sch{\"o}lkopf.
\newblock Learning with local and global consistency.
\newblock In \emph{Advances in neural information processing systems}, pages
  321--328, 2004.

\bibitem[Zhu et~al.(2005)Zhu, Lafferty, and Rosenfeld]{zhu2005semi}
Xiaojin Zhu, John Lafferty, and Ronald Rosenfeld.
\newblock \emph{Semi-supervised learning with graphs}.
\newblock PhD thesis, Carnegie Mellon University, language technologies
  institute, school of~…, 2005.

\end{thebibliography}

\newpage

\begin{appendices}

\title{Supplementary Material - Contrastive Mixup: Self- and Semi-Supervised learning for Tabular Domain  }
\maketitle

\section{Contrastive Mixup}
The proposed pre-training is summarized in Algorithm. \ref{alg:proposed-method}. The encoder is initially trained by using the labeled subset for the contrastive loss component Eqn. 6, and both unlabeled \& labeled subsets for the reconstruction loss Eqn.7. Subsequently, after $K$ epochs via label propagation we generate pseudo labels for the unlabeled subset so they can be leveraged in the contrastive loss as well.

\begin{algorithm}[H]
\label{alg:proposed-method}
\caption{the proposed method's main algorithm.}
 \hspace*{\algorithmicindent} \textbf{Input:} constant $\tau$, encoder $e$, decoder $f$, projection network $h^p$, labeled set $(x_l, y_l)$, unlabeled set with pseudolabels if available $(x_u, y_{u})$ 
    \begin{algorithmic}[1]
        \For{sampled mini-batches $B_L, B_u$ from $\{(x_l, y_l)\}_{l=1}^{N_l}, \{(x_u, y_{u})\}_{u=1}^{N_U}}$
        \State draw $\lambda \sim \text{Uniform}(0, \alpha)$ of size $B_l + B_u$ \Comment{$\alpha \in [0.0, 0.5]$}
        \State draw random integer $i \in [0, \mathcal{I}]$ 
        \Comment{$\mathcal{I} :=$ number of layers in encoder}
        \State $h_l^i = e_{0:i}(x_l)$, ~~ $h_u^i = e_{0:i}(x_l)$ 
        \Comment{Feed through $0$ to $i^{th}$ layer of encoder}
        \State $h_{\text{mixed}}^i =$ Mixup$([h_l^i; h_u^i], [y_l; y_u], \lambda)$ \Comment{Mix within the same label}
        \State $z_l = e_{i:\mathcal{I}}(h_l^i)$
        \State $z_u = e_{i:\mathcal{I}}(h_u^i)$
        \State $z_{\text{mixed}} = e_{i:\mathcal{I}}(h_{\text{mixed}}^i)$
        \State $l_{\text{recon}} = l_r([x_L; x_U])$ \Comment{Eqn. 7}
        \State $h_l^p = h^p(z_l)$ 
        \State $h_u^p = h^p(z_u)$
        \State $h_{\text{mixed}}^p = h^p(z_{\text{mixed}})$
        \State $l_{\text{contrastive}} = l^{sup}_{\tau}([h_l^p;h_u^p], h_{\text{mixed}}^p)$ \Comment{$l^{sup}(view 1, view 2)$ Eqn. 6}
        \State $\mathcal{L} = l_{\text{contrastive}} + l_{recon}$
        \State Update networks $e, f, $ and $h^p$ to minimize $\mathcal{L}$
        \EndFor
    \end{algorithmic}
\end{algorithm}

\subsection{Limitations}
Underlying our method, we make use of the \textit{Manifold Assumption} where high dimensional data lies (roughly) on a lower-dimensional manifold to avoid creating low probable samples through interpolation in the original data space. Further, as the data manifold may change for different labelled examples, in our method, we enforce mixing within the same labelled class, limiting the set of labelled samples that are initially used in the contrastive component. To leverage the unlabeled subsets, we generate pseudo-labels for the unlabeled samples to be used in the contrastive loss component. This makes the method more reliable on the quality of the initially labelled subset of examples. Additionally, as the method relies on discrete labels, in its current presentation cannot be applied to regression tasks.


\section{Additional Experiments}
\subsection{Mixing within class vs randomly}
We compare limiting Mixup augmentation in the latent space to same labeled samples versus the original Mixup augmentation where any random set of samples can be used to interpolate in between to generate new samples. When randomly interpolating between samples, $\tilde{h} = \lambda h_1 + ( 1-\lambda)h_2$ we enforce $h_1$ to be $\lambda$ close to $\tilde{h}$ and $h_2$ $1-\lambda$ close to $\tilde{h}$ in the contrastive term. In this experiment the Mixup component for the contrastive loss was changed and the rest is untouched. As can be seen from Table. \ref{tab:randommixingvsclassmixing} mixing between random samples in the latent space can hurt performance compared to mixing within the same labeled class.

\begin{table*}[h]%
\renewcommand{\arraystretch}{1.2}
\caption{\label{tab:randommixingvsclassmixing} Comparison on between randomly interpolating between examples and interpolating within the same labeled class examples.}
\begin{minipage}{\textwidth}
\begin{center}
\resizebox{\columnwidth}{!}{
\begin{tabular}{ll | ccccc}
\toprule
& &  \multicolumn{4}{c}{\textbf{Dataset}} \\
\textbf{Type} & \textbf{Method} & \textbf{MNIST} & \textbf{Adult} & \textbf{Blog Feedback} & \textbf{Covertype} \\
\hline
\multirow{1}{*}{\textbf{Ours (Ablation)}}
& Random Mixing & {96.60} {\scriptsize($\pm 0.111$)}& {84.33} {\scriptsize($\pm 0.472$)}& {81.72} {\scriptsize($\pm 0.389$)} & {79.71} {\scriptsize($\pm 0.229$)}\\
&  Ours & \textbf{{97.58} {\scriptsize($\pm 0.078$)}}& \textbf{85.42} {\scriptsize($\pm 0.210$)}& \textbf{81.88} {\scriptsize($\pm 0.123$)}& \textbf{80.41} {\scriptsize($\pm 0.205$)}\\
\hline
\end{tabular}}
\end{center}
\end{minipage}
\end{table*}

\subsection{MNIST Limited Samples}
\begin{figure}
    \centering
    \includegraphics[scale=0.75]{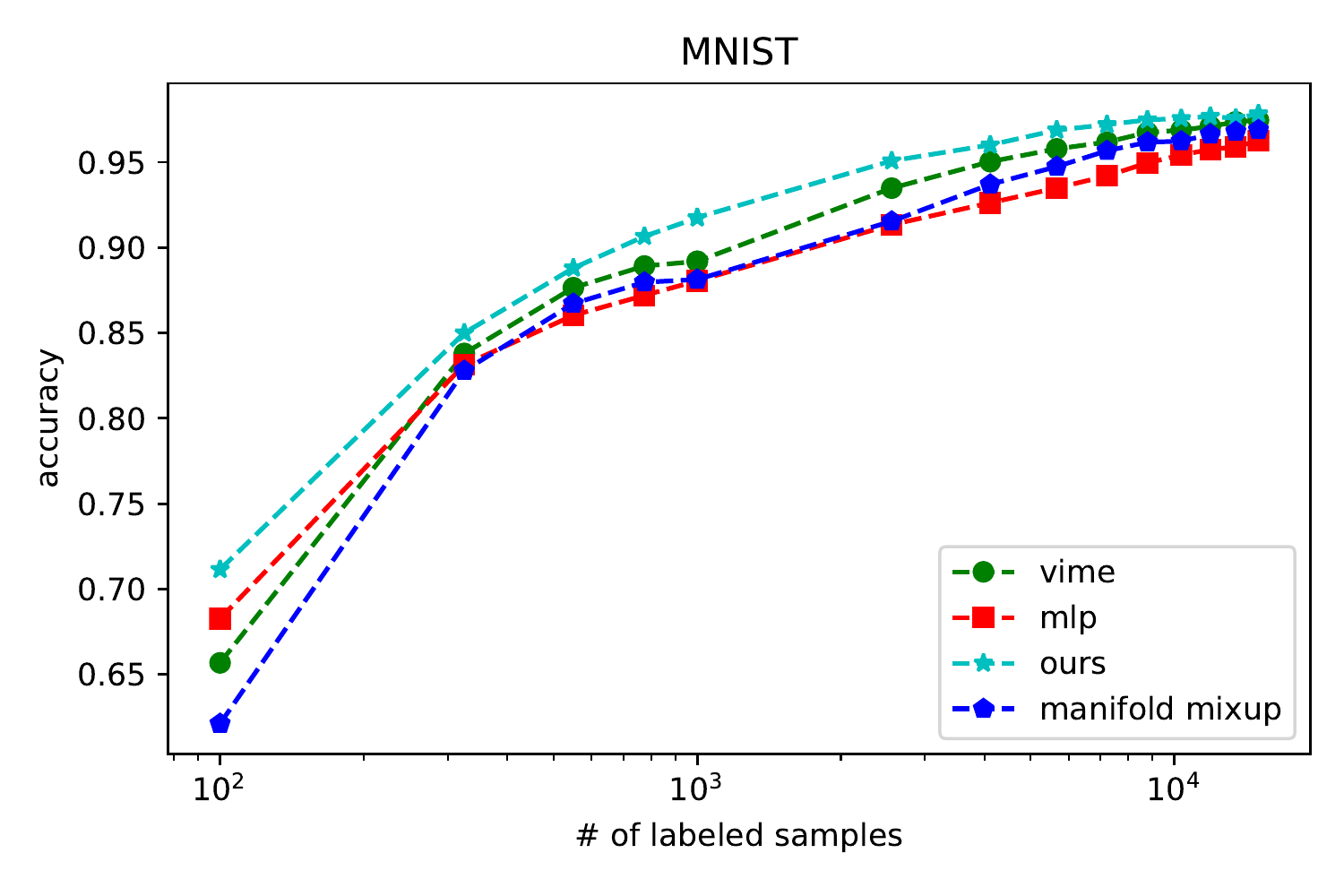}
    \caption{Comparison of accuracy performance on MNIST under varying number labeled examples used for training.}
    \label{fig:mnist_label_curve}
\end{figure}

In Figure. \ref{fig:mnist_label_curve} we run an additional experiment to demonstrate the effectiveness of the proposed method under limited number of labeled samples. The proposed framework consistently outperforms baselines.

\begin{figure}
    \centering
    \includegraphics[scale=0.75]{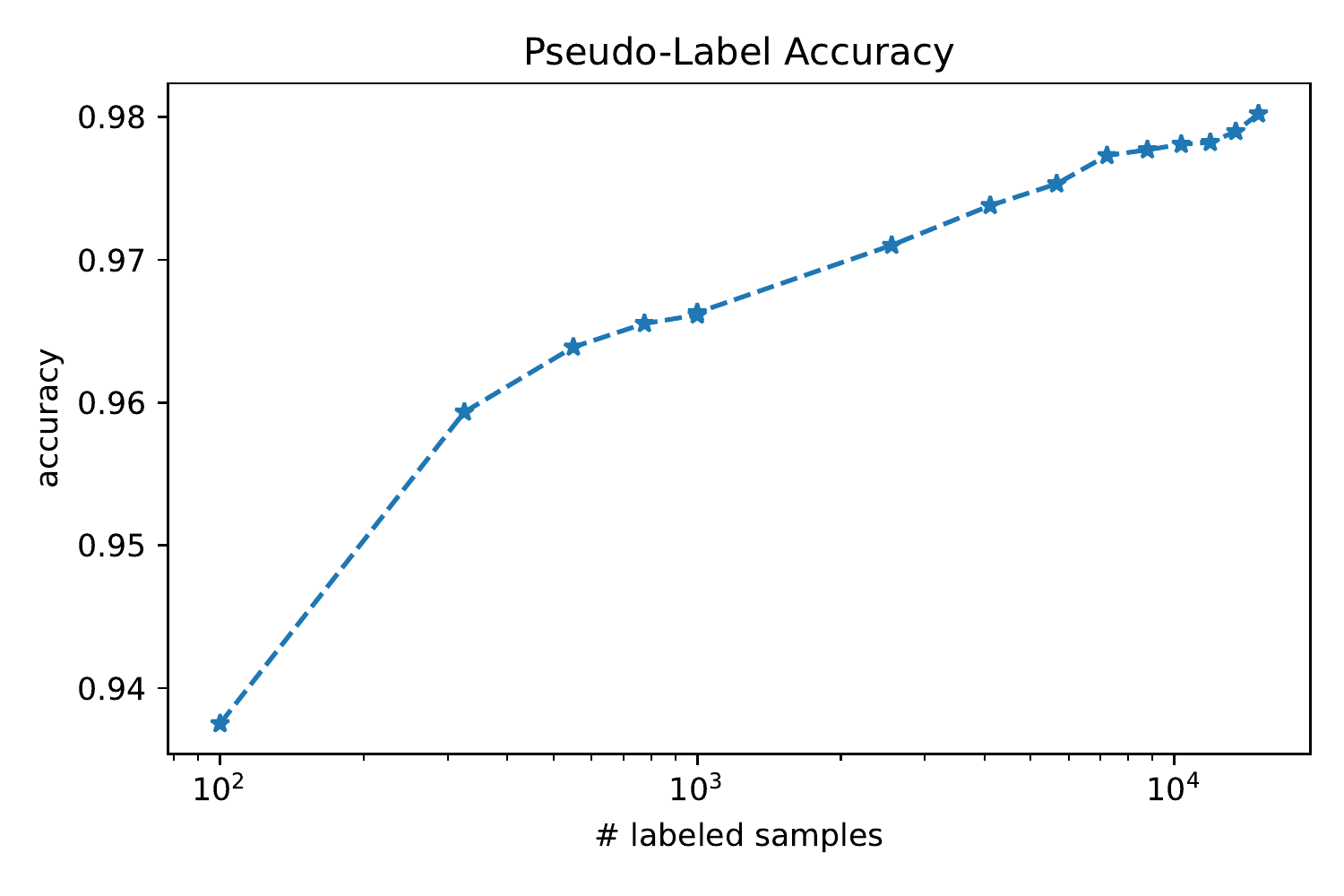}
    \caption{MNIST pseudo-labeling accuracy across varying number of labeled samples. Accuracy is reported after training for 20 epochs.}
    \label{fig:pseudo-labeling}
\end{figure}

In Figure. \ref{fig:pseudo-labeling} we conduct an experiment to evaluate the pseudo-labeling accuracy as a function of number of labeled samples projected in the latent space.  

\begin{figure}
    \centering
    \includegraphics[scale=0.75]{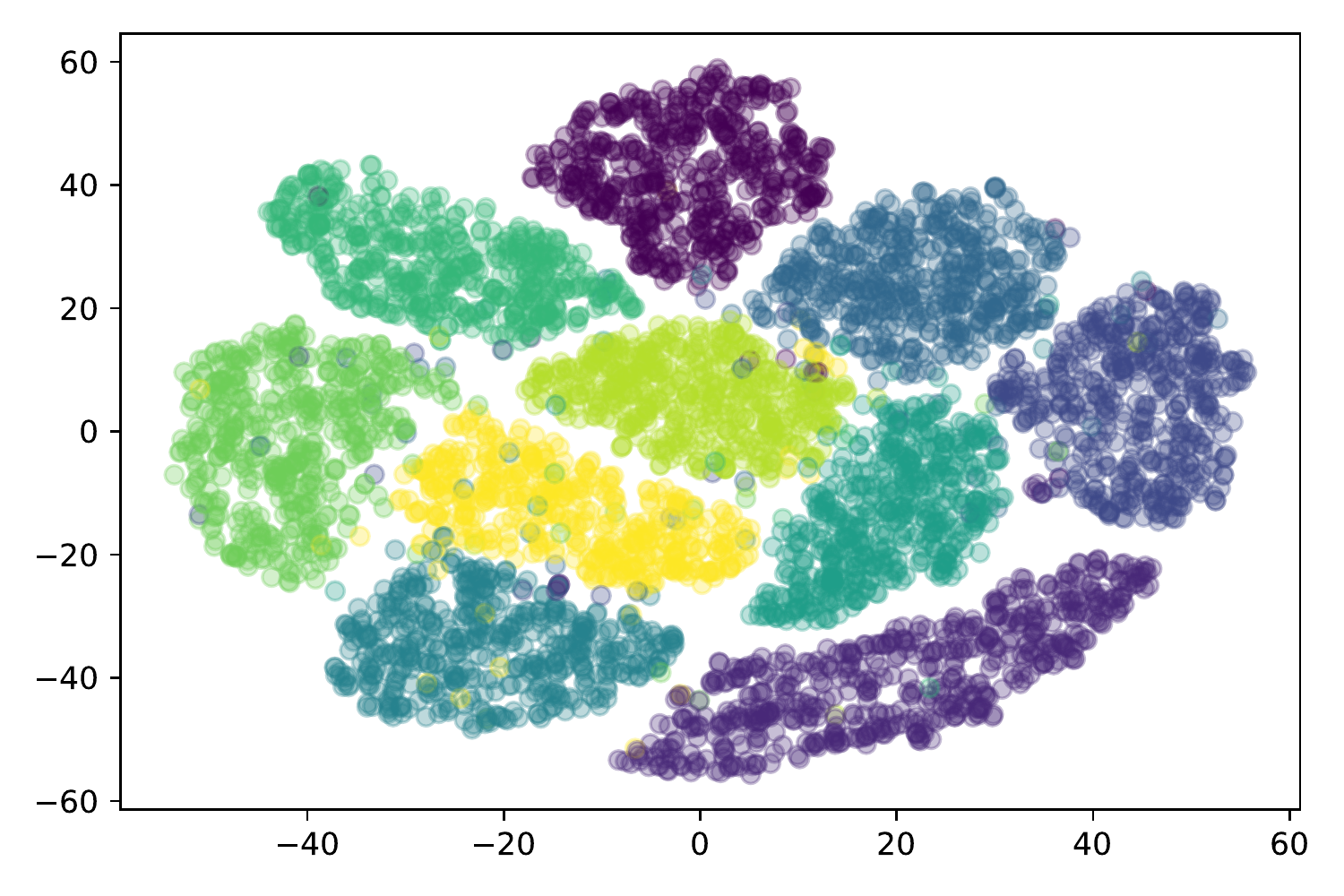}
    \caption{TSNE visualization of representations extracted from encoder training on 10\% labeled examples of MNIST.}
    \label{fig:genomics_lr_acc_curve}
\end{figure}

\label{sec:implementation-details}
\section{Implementation Details}
Our proposed framework consists of 5 different components: (1) encoder, (2) decoder, (3) Within Class Mixup + Contrastive Loss, (4) Label propagation (5) Predictor.

\begin{enumerate}
    \item Our encoder is a set of fully connected layers, where the number of hidden layers and the number of layers is a hyper-parameter.
    \item A weight hyperparameter $\beta$ is used to control the decoder loss term. The decoder has the same architecture as the encoder, but in reverse.
    \item The proposed within class Mixup has a hyperparameter $\alpha \in [0.0, 0.5]$ that is used to sample $\lambda$ from $\text{Uniform}(0, \alpha)$. The projection network is a set of fully connected layers with projection dimension $d_{proj}$, and number of layers $L_{proj}$.
    \item The label propagation component contains hyperparameters $k$ for the number of nearest neighbors and $\alpha$ parameter. $k=3$ and $\alpha=0.999$ is used as the default across all experiments.
    \item The predictor is a set of fully connected layers. \emph{(FC-BN-ReLU)}. 
\end{enumerate}

Experiments were run on two sets of datasets. On public datases we set the hyperparameters as follows:
\begin{enumerate}
    \item Hidden layer dimension size set to be the same as $d$ after embedding categorical columns. The embeddings generated for each categorical column is set to $min(600, round(1.6 * d_{|\mathcal{D}_i|}^{0.56}))$ and the number of layers is $1$.
    \item Decoder reconstruction loss weight $\beta=0.25$
    \item $alpha=0.2$ for within class Mixup, $L_{proj}=1$, and $d_{proj}=d$ 
    \item default parameters
    \item Our predictor is a $2$ layer MLP with hidden dimension size of $100$. Mixup augmentation is set to default setting with $\alpha=1.0$ and $\lambda \sim \text{Uniform}(0, \alpha)$
\end{enumerate}

On the genomics dataset we tune hyperparameters using the  \href{https://github.com/microsoft/nni}{Neural Network Intelligence (NNI) Auto-Tuner} and pick the best hyper-parameter setting on validation.

\section{Experimental Details}
We evaluated the performance of our method on 4 public datasets and 4 phenotypes. In each of these datasets $10\%$ of the samples in the training set was used as labeled and the reset as unlabeled. This labeled subset depends on the random seed in our implementation, and a total of 5 experiments was run for each dataset by varying the random seed $[123, 127, 131, 137, 130]$ and report the average following hyperparameter selection.

On public datasets, the difference between VIME and our proposed framework is purely on the training algorithm, and the same capacity networks are used through out i.e. in VIME 4 networks are used encoder, feature estimator, mask estimator, predictor, and these are the same as our encoder, decoder, projector, predictor respectively.

On genomics dataset, the different methods were tuned on validation using \href{https://github.com/microsoft/nni}{Neural Network Intelligence (NNI) Auto-Tuner}.

\section{Dataset Details}
\subsection{UK Biobank}
\begin{minipage}{\textwidth}
\begin{center}
\begin{tabular}{cccccc}
 \toprule
 ID & Dataset  & Categorical & Continuous & Num Samples\\
 \midrule
 1 & MPV & 714 & 0 & 2913273 \\
 \midrule
 2 & Smoking Status & 714 & 0 & 2913273\\
 \midrule
 3 & MSCV & 1950 & 0 & 2913273\\
 \midrule
 4 & Hair Color & 1000 & 0  & 2913273\\
 \bottomrule
\end{tabular}
\end{center}
\end{minipage}

\subsection{Public Datasets}

The public datasets used are summarized in Table \ref{tab:pub-data-stats}. Three of the datsets 1, 2, 3 contain separate test sets, and covtype (4) 20\% of the data is used as test and the rest for train. For each dataset we use $10\%$ of the train set as labeled and the rest as unlabeled.

\begin{minipage}{\textwidth}
\begin{center}
\label{tab:pub-data-stats}
\begin{tabular}{cccccc}
 \toprule
 ID & Dataset  & Categorical & Continuous & Num Samples\\
 \midrule
 1 & \href{http://yann.lecun.com/exdb/mnist/}{mnist} & 0 & 724 & 60000\\
 \midrule
 2 & \href{https://archive.ics.uci.edu/ml/datasets/census+income}{adult} & 8 & 6 & 48840\\
 \midrule
 3 & \href{https://archive.ics.uci.edu/ml/datasets/BlogFeedback}{BlogFeedback} & 213 & 67 & 60021\\
 \midrule
 4 & \href{https://archive.ics.uci.edu/ml/datasets/covertype}{covtype}& 44 & 10  & 581011\\
 \bottomrule
\end{tabular}
\end{center}
\end{minipage}

\section{Software \& Hardware}
Experiments were run on a machine with a GeForce RTX 2080 TI, 128 Gb RAM, and Intel(R) Core(TM) i9-7920X CPU.

To ensure reproducibility, all experiments were run using the same set of random seeds, baseline methods are re-implemented in the same code base and the same software versions are used.
\begin{table}[h]%
\renewcommand{\arraystretch}{1.3}
\caption{Python dependencies.}
\begin{minipage}{\textwidth}
\begin{center}
\texttt{
\begin{tabular}{l|l}
\toprule
\textbf{Dependency} & \textbf{Version} \\
\hline
python & 3.6.1 \\
pytorch & 1.7.1 \\
numpy & 1.15.4 \\
pandas & 1.1.5 \\
scikit-learn & 0.24.2 \\
scipy & 1.6.3 \\
tqdm & 4.60 \\
matplotlib & 3.4.1 \\
\bottomrule
\end{tabular}}
\end{center}
\end{minipage}
\label{tab:swdeps}
\end{table}%

\end{appendices}

\end{document}